\documentclass[runningheads]{llncs}

\usepackage[mobile]{eccv}

\usepackage{eccvabbrv}

\usepackage{graphicx}
\usepackage{booktabs}

\usepackage[accsupp]{axessibility}  

\usepackage{hyperref}

\usepackage{orcidlink}

\begin{document}

\title{First Place Solution to the ECCV 2024 ROAD++ Challenge @ ROAD++ Atomic Activity Recognition 2024} 


\author{Ruyang Li \and
Tengfei Zhang \and
Heng Zhang \and
Tiejun Liu \and
\\
Yanwei Wang \and
Xuelei Li
}

\authorrunning{R. Li et al.}

\institute{Inspur (Beijing) Electronic Information Industry Co., Ltd., Beijing, P.R.China
\email{\{liruyang,zhangtengfei,zhangheng10,liutj,wangyanwei,lixuelei\}@ieisystem.com}
}

\maketitle

\begin{abstract}
    This report presents our team's technical solution for participating in Track 3 of the 2024 ECCV ROAD++ Challenge. The task of Track 3 is atomic activity recognition, which aims to identify 64 types of atomic activities in road scenes based on video content.
    Our approach primarily addresses the challenges of small objects,
    discriminating between single object and a group of objects, as well as model overfitting in this task. 
    Firstly, we construct a multi-branch activity recognition framework that not only separates different object categories but also the tasks of single object and object group recognition, thereby enhancing recognition accuracy. 
    Subsequently, we develop various model ensembling strategies, including integrations of multiple frame sampling sequences, different frame sampling sequence lengths, multiple training epochs, and different backbone networks.
    Furthermore, we propose an atomic activity recognition data augmentation method, which greatly expands the sample space by flipping video frames and road topology, effectively mitigating model overfitting.
    Our methods rank first in the test set of Track 3 for the ROAD++ Challenge 2024, and achieve 69\% mAP.
  \keywords{Atomic activity recognition \and Action recognition \and Video classification}
\end{abstract}

\section{Introduction}
\label{sec:intro}
Precise detection and identification of road participants, such as pedestrians, vehicles, bicycles, and others, are crucial for the safety of autonomous driving vehicles. 
The outcomes of these detection and identification efforts are key to bolstering the decision-making abilities of autonomous cars. 
As such, the ECCV 2024 ROAD++ Challenge~\footnote{https://sites.google.com/view/road-eccv2024/challenge}~\cite{singh2022road} aims to explore the development of semantically rich representations of road scenes based on the concept of road events, all in the pursuit of advancing autonomous driving technology. 
The challenge is divided into three tracks: spatiotemporal agent detection, spatiotemporal road event detection, and multi-label atomic activity recognition~\cite{kung2023action}. This paper is about the last track.

\begin{figure}[htbp]
  \centering
  \includegraphics[width=\linewidth]{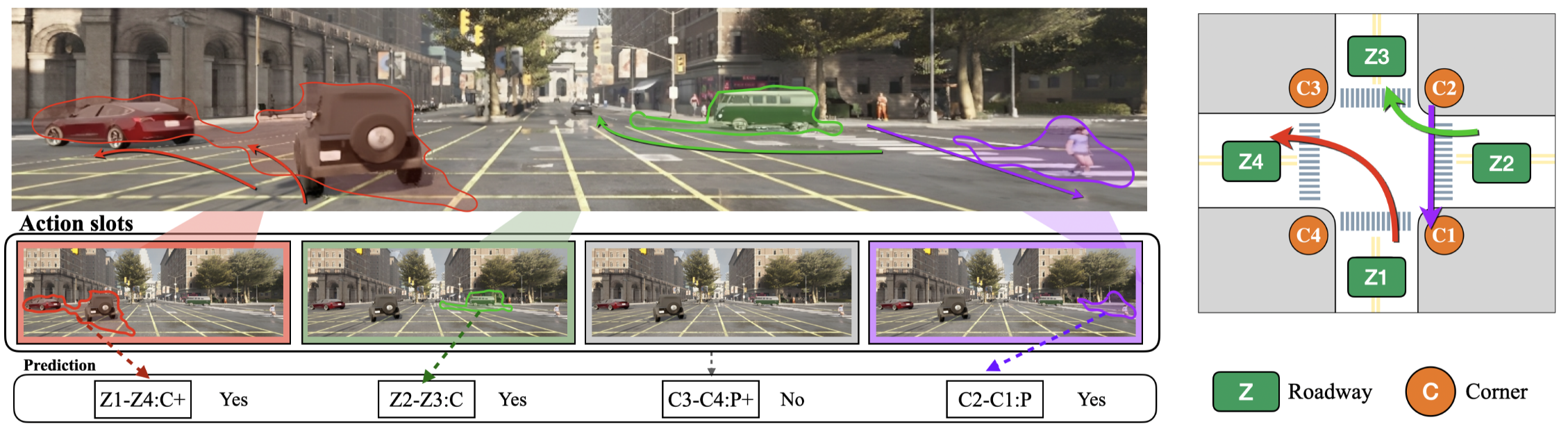}
  \caption{Illustration of multi-label atomic activity recognition~\cite{kung2023action}.
  }
  \label{fig:task_intro}
\end{figure}

The objective of the third task is to achieve multi-label atomic activity recognition, also known as multi-label action recognition, which involves identifying 64 types of atomic activities in road scenes based on video content. 
Atomic activity classes are defined as (region\_start $\rightarrow$ region\_end: agent\_type), where region\_start and region\_end represent the road topology at intersections, and agent\_type refers to the type of agent (object), such as vehicles or pedestrians. 
As shown in~\cref{fig:task_intro}, for an intersection, Z1 to Z4 represent four roads, and C1 to C4 represent four pedestrian waiting areas. 
C and C+ denote a single vehicle and a group of vehicles, respectively, while P and P+ denote a single pedestrian and a group of pedestrians, respectively. 
Based on these definitions, an atomic activity can be easily represented. 
For instance, the red arrow in~\cref{fig:task_intro} can be denoted as \textbf{Z1-Z4:C+}, which signifies a group of vehicles traveling from road Z1 to road Z4.

We conduct an in-depth analysis of the challenges in this task. 
Firstly, some small objects, such as pedestrians at a distance, contain very few pixels and are therefore easily overlooked. 
Additionally, distinguishing between single object and object groups is quite challenging. 
On one hand, when objects are small, it is difficult for the model to differentiate between single object and object groups.
On the other hand, when objects are in close proximity, it can lead to incorrect identification of single object and object groups. 
The last challenge is model overfitting. The data for this task is collected from a simulated environment, which, compared to real-world data, is overly regular or uniform in some characteristics. This makes the model prone to overfitting and increases the difficulty of model optimization.

To solve the above challenges, we propose a series of solutions. 
Firstly, we construct a multi-branch atomic activity recognition framework. It not only separates different object categories but also distinguishes the recognition tasks of single object and object groups, thereby mitigating the difficulty of recognizing individual object and object groups. 
Subsequently, we introduce various model ensembling strategies, including the integration of multi-frame sampling sequences, integration of different frame sampling sequence lengths, integration across multiple training epochs, and integration of different backbone networks. 
The integration of frame sampling sequences enhances the coverage of video content. 
The integrations across multiple training epochs and different backbone networks combine the strengths of various models to further improve recognition accuracy. 
Additionally, we propose an atomic activity recognition data augmentation method that greatly expands the sample space by flipping video frames and road topology, effectively alleviating model overfitting.

\section{Challenge Analysis}
We identify some challenges in this task, including small objects, distinguishing between single object and object groups, and model overfitting.

\subsection{Small Objects}
\begin{figure}[htbp]
  \centering
  \includegraphics[width=\linewidth]{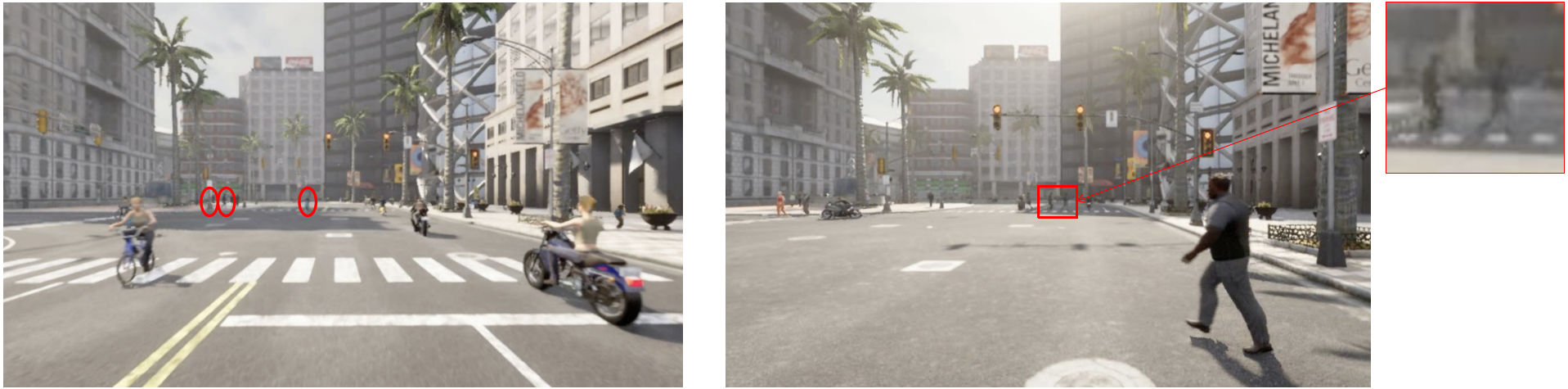}
  \caption{Examples of small objects.
  }
  \label{fig:small_object}
\end{figure}

As shown in the left image of~\cref{fig:small_object},
small objects, such as distant pedestrians, have very few pixels and can easily be overlooked, limiting the recognition accuracy.
Besides, small objects may lead to incorrect classification of single object and object groups. 
As shown in the right image of~\cref{fig:small_object},
multiple adjacent small objects may be judged as a single object by the model.

\subsection{Single Object and Object Group Recognition}
The task of distinguishing between individual objects and groups of objects presents a challenge.
Beyond the interference from small objects mentioned above,
even when the objects are not small, the proximity of objects to each other makes it difficult to differentiate them on high-level feature maps.
Furthermore, the absence of object detection task increases the difficulty in classify whether an entity is a single object or an object group.

\begin{figure}[htbp]
  \centering
  \begin{subfigure}{0.3\linewidth}
    \includegraphics[width=\linewidth]{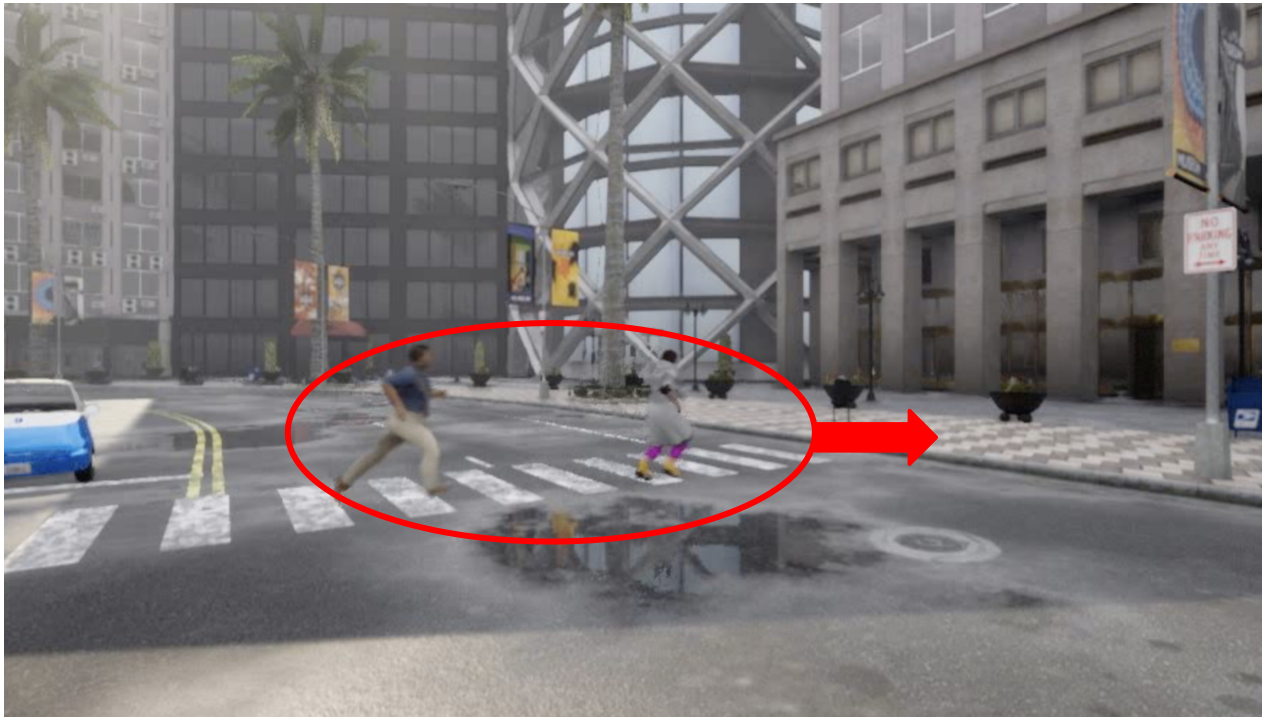}
    \caption{}
    \label{fig:challenge3_01}
  \end{subfigure}
  \hfill
  \begin{subfigure}{0.3\linewidth}
    \includegraphics[width=\linewidth]{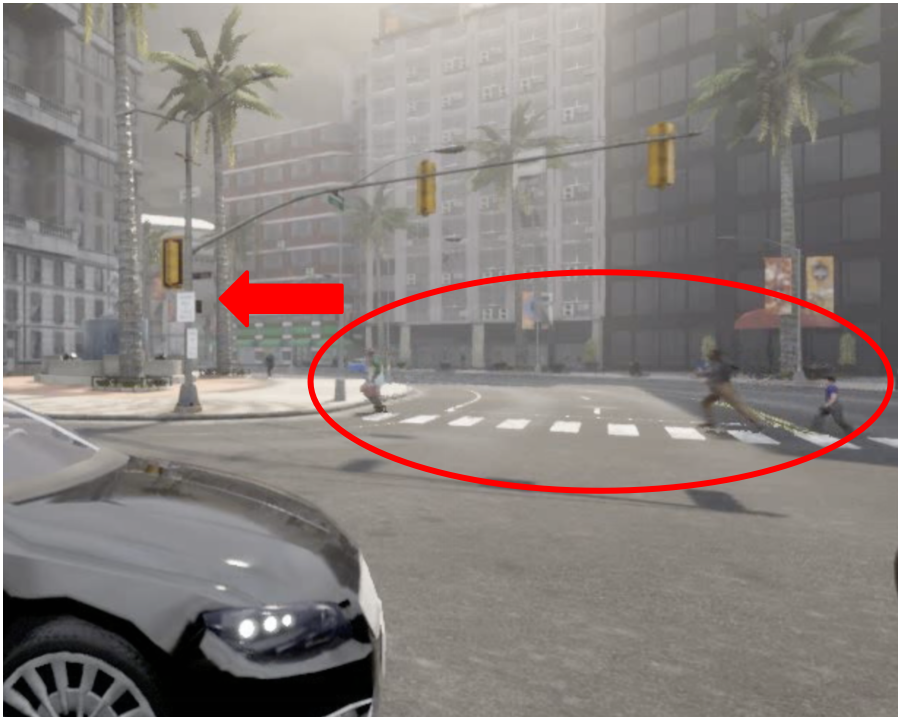}
    \caption{}
    \label{fig:challenge3_02}
  \end{subfigure}
    \hfill
  \begin{subfigure}{0.3\linewidth}
    \includegraphics[width=\linewidth]{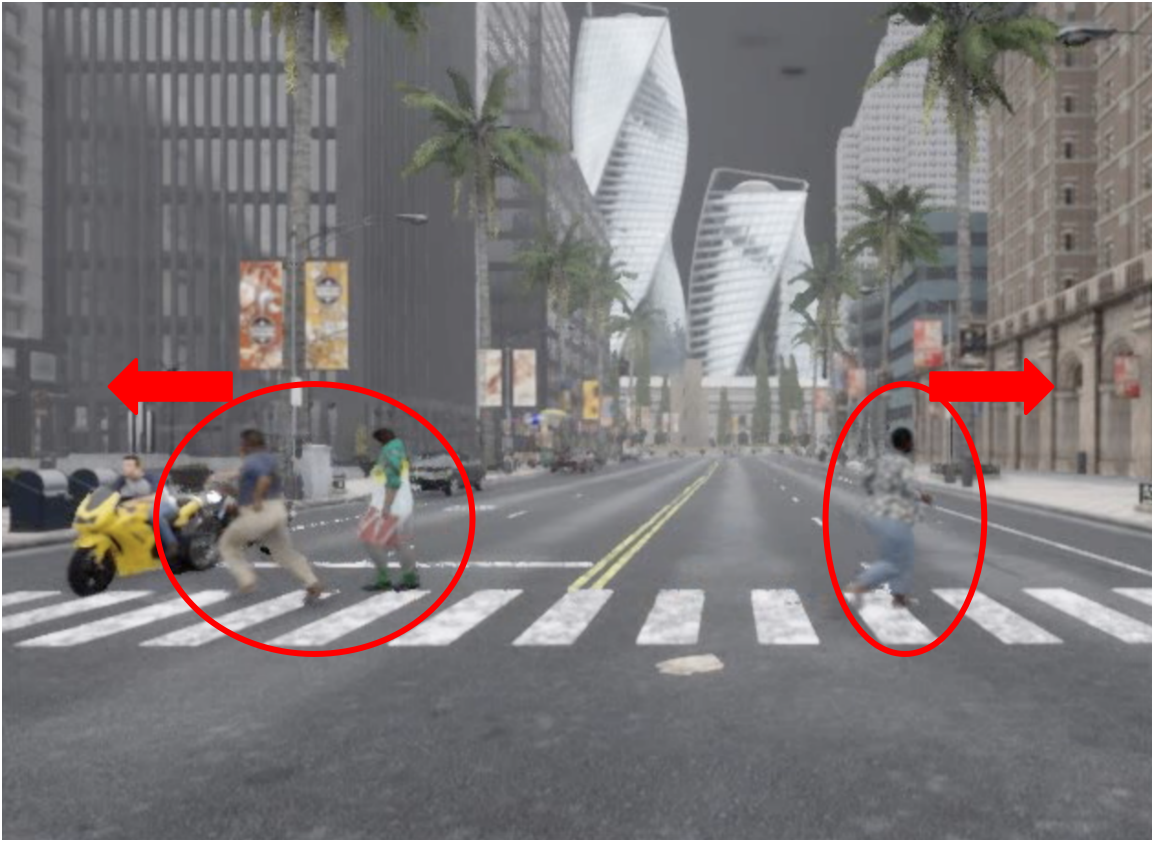}
    \caption{}
    \label{fig:challenge3_03}
  \end{subfigure}
  \caption{Examples of single objects and object groups.}
  \label{fig:challenge3}
\end{figure}

\cref{fig:challenge3} provides some examples of individual pedestrians and groups of pedestrians. 
Due to the large receptive field of deep networks, some objects that are close to each other may be merged into a single feature. 
Additionally, if the individuals within a group of objects are too far apart, it may also cause the model to fail to classify these objects into a group.

\subsection{Model Overfitting}
Overfitting is a common issue. 
This task involves data collected from a simulation environment. 
Compared to real-world data, it may be too regular or uniform in some features, leading to model overfitting. 
This makes it challenging to select the optimal model.

\section{Solutions}
We develop solutions to address the above challenges, which include a multi-branch atomic activity recognition framework, model ensembling, and video data augmentation. Our solutions are based on the codebase of Action-slot~\cite{kung2023action}.

\subsection{Multi-Branch Framework}
\begin{figure}[htbp]
  \centering
  \includegraphics[width=0.6\linewidth]{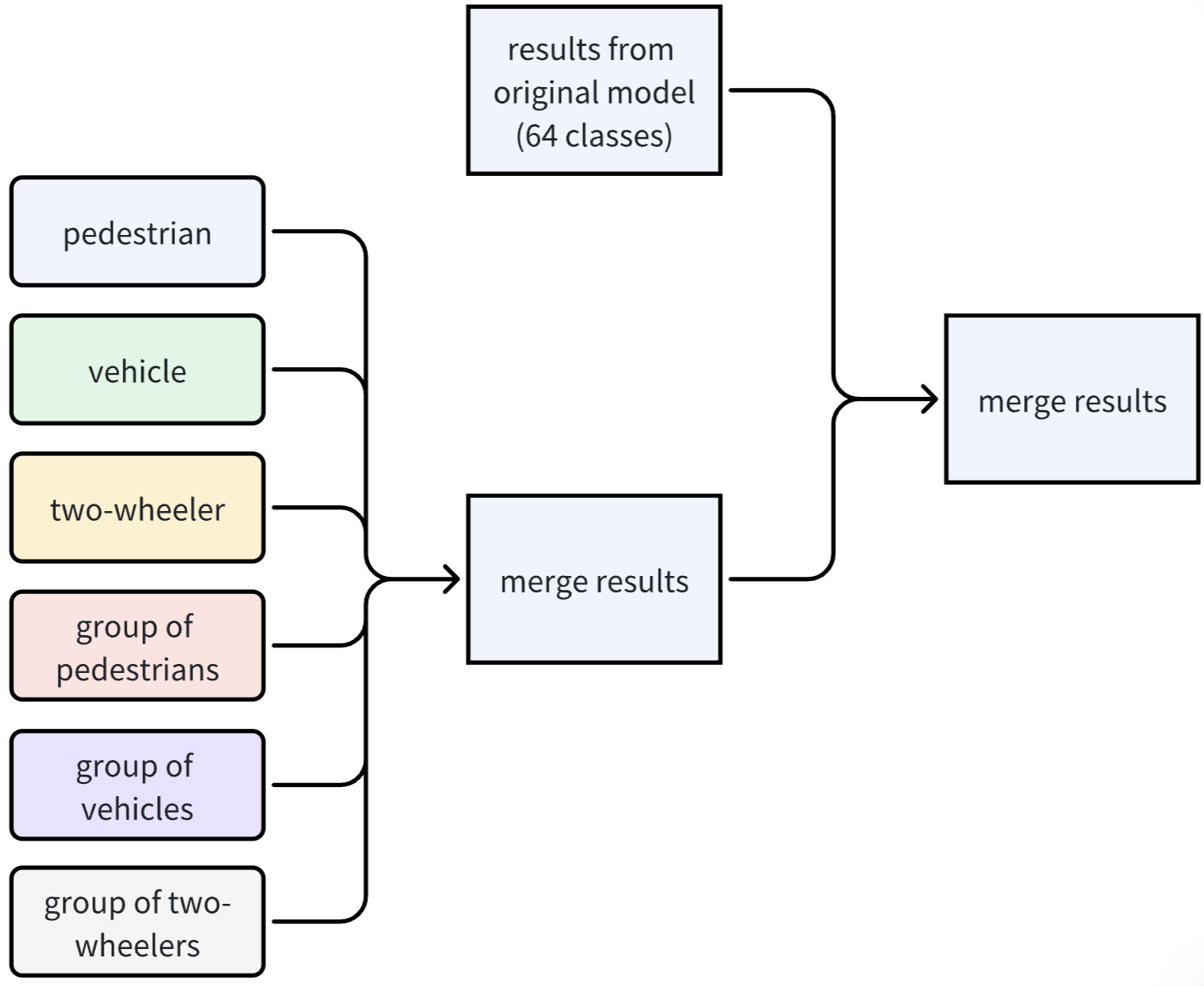}
  \caption{The multi-branch atomic activity recognition framework.
  }
  \label{fig:multi_branch}
\end{figure}

For the problem of classifying objects and object groups, we develop a multi-branch atomic activity recognition framework.
As shown in~\cref{fig:multi_branch}, this framework separates the recognition tasks of objects and object groups, processing them with different models respectively. 
Then, the classification results are merged.
Finally, we combine the merged results with the prediction results of the standard model.
The benefit of this approach is clear, each model only needs to concentrate on actions within a specific category, which simplifies the learning process.

\subsection{Model Ensembling}
We propose a series of model ensembling methods, including the integration of multi-frame sampling sequences, integration of different frame sampling sequence lengths, integration across multiple training epochs, and integration of different backbone networks. 

\subsubsection{1) Integration of prediction results from multiple frame sampling sequences}
\begin{figure}[htbp]
  \centering
  \includegraphics[width=\linewidth]{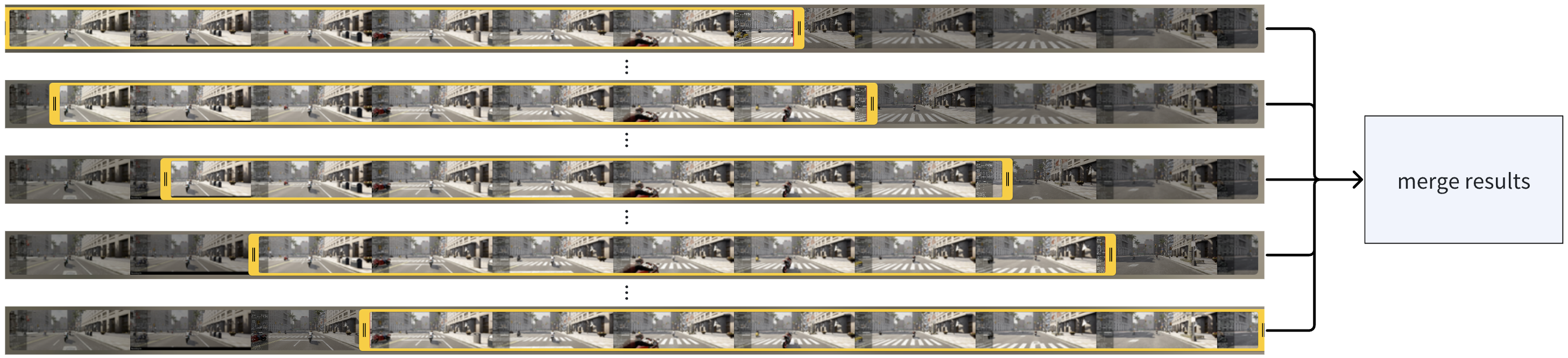}
  \caption{Integration of prediction results from multiple frame sampling sequences.
  }
  \label{fig:fuse_multi_frame_sequences}
\end{figure}

We propose the integration of prediction results from multiple frame sampling sequences, as shown in~\cref{fig:fuse_multi_frame_sequences}.
This method is only employed during the model testing phase.
During testing, unlike the baseline approach~\cite{kung2023action} of only selecting the middle frame sampling sequence, we use all frame sampling sequences for prediction and merge all the prediction results.

\subsubsection{2) Integration of prediction results from different frame sampling sequence lengths}
\begin{figure}[htbp]
  \centering
  \includegraphics[width=0.4\linewidth]{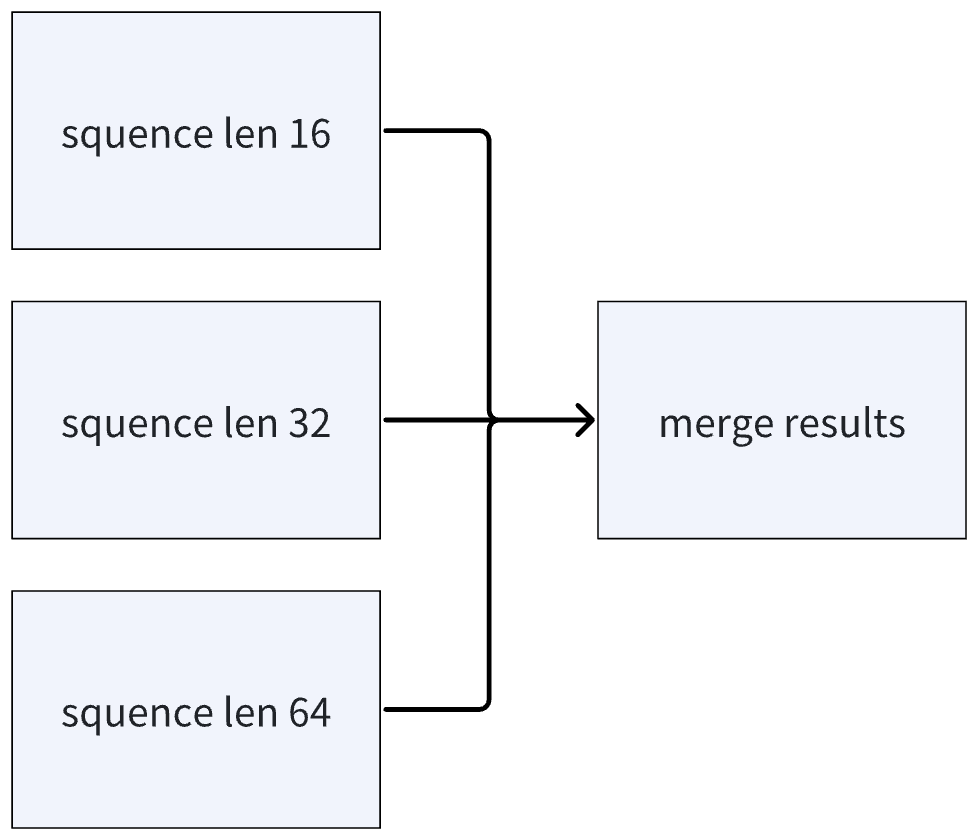}
  \caption{Integration of prediction results from different frame sampling sequence lengths.
  }
  \label{fig:fuse_sequences_length}
\end{figure}

The second model ensembling method is the integration of prediction results from different frame sampling sequence lengths.
That is, we train models using different lengths of frame sampling sequences and combine the prediction results from these models, as illustrated in~\cref{fig:fuse_sequences_length}.

\subsubsection{3) Integration of prediction results from multiple training epochs}
\begin{figure}[tb]
  \centering
  \includegraphics[width=0.4\linewidth]{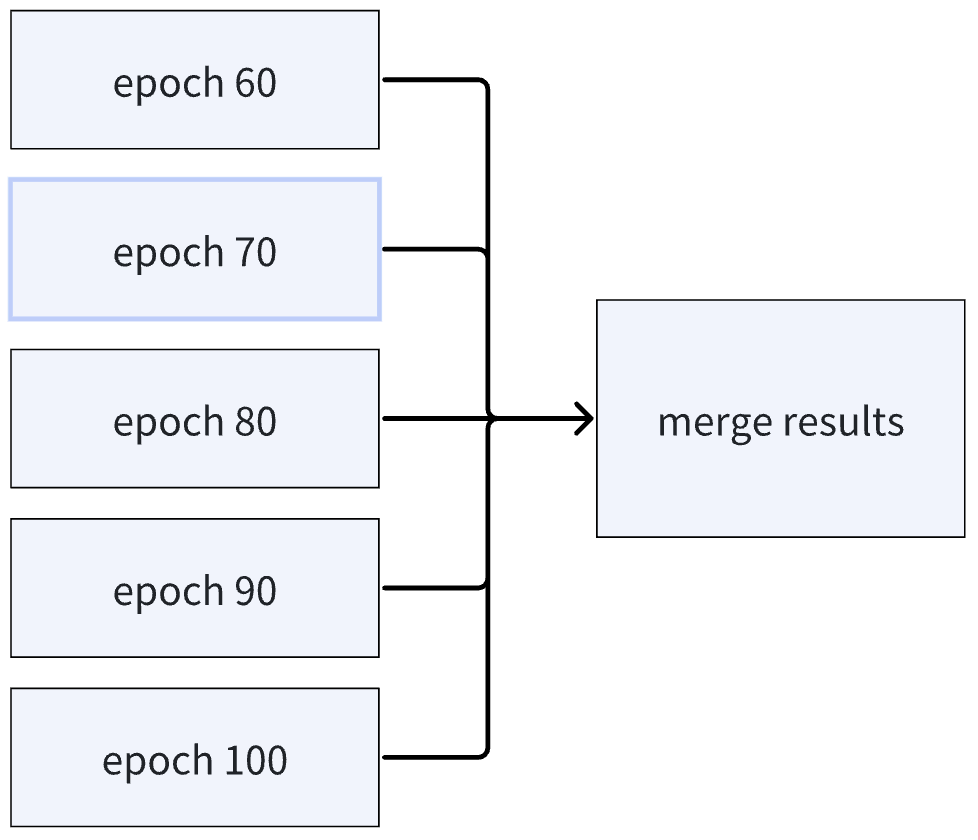}
  \caption{Integration of prediction results from multiple training epochs.
  }
  \label{fig:fuse_multi_epochs}
\end{figure}

As shown in~\cref{fig:fuse_multi_epochs}, the third model ensembling method is the integration of prediction results from multiple training epochs. As mentioned earlier, overfitting makes it difficult to select the optimal model, and we find that the best model should be between the 60 and 100 epochs. Therefore, we integrate the prediction results of these models.

\subsubsection{4) Integration of models with different backbone networks}
The last model ensemble method is the integration of models with different backbone networks. 
X3D~\cite{X3D} and InternVideo~\cite{wang2022internvideo} are employed as backbone networks.
InternVideo is a general video foundation model via generative and discriminative learning.

\subsection{Data Augmentation}
\begin{figure}[tb]
  \centering
  \includegraphics[width=0.6\linewidth]{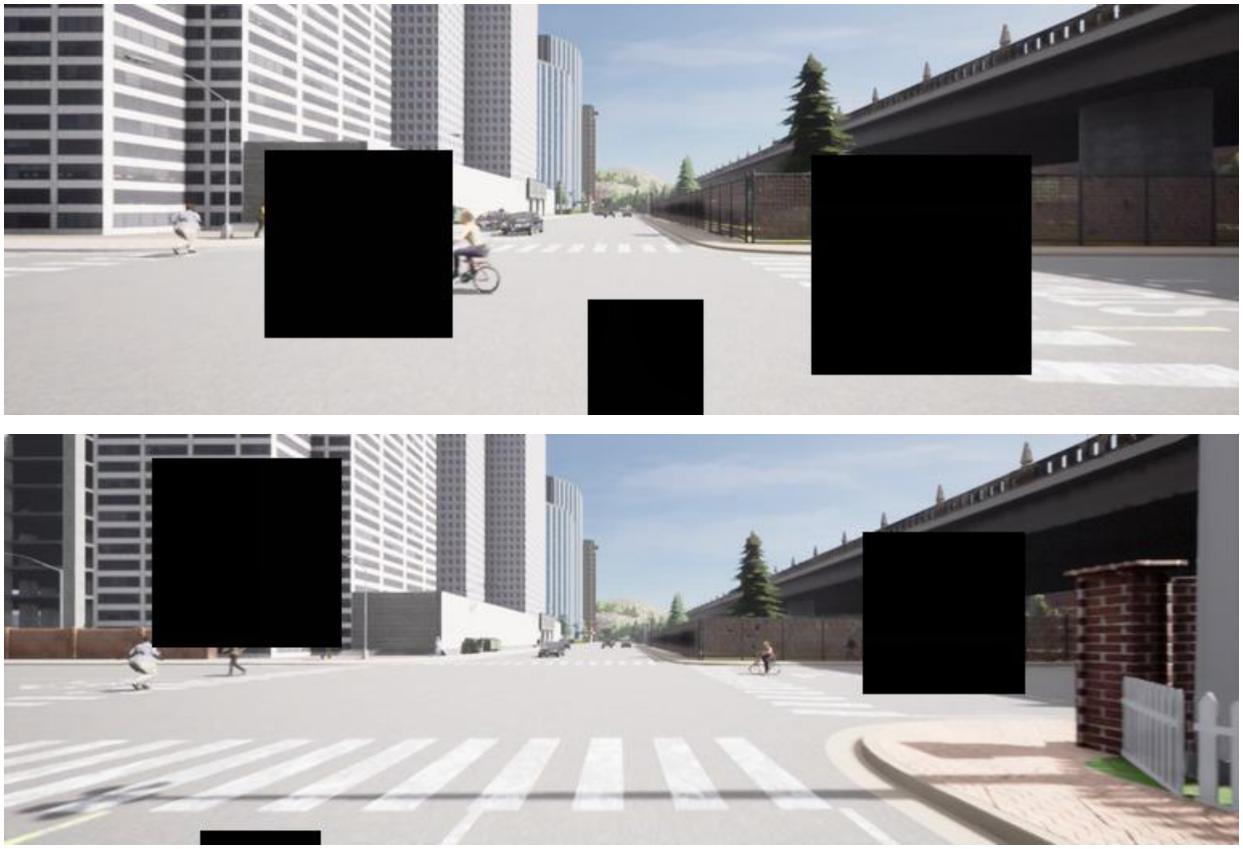}
  \caption{Examlpe of Cutout operation.
  }
  \label{fig:cutout}
\end{figure}

We employ data augmentation to mitigate the overfitting issue mentioned above.
We introduce two data augmentation methods. 
The first data augmentation method is Cutout~\cite{devries2017improved}, which randomly covers a region of an input image with a square, as shown in~\cref{fig:cutout}.

\begin{figure}[tb]
  \centering
  \includegraphics[width=0.8\linewidth]{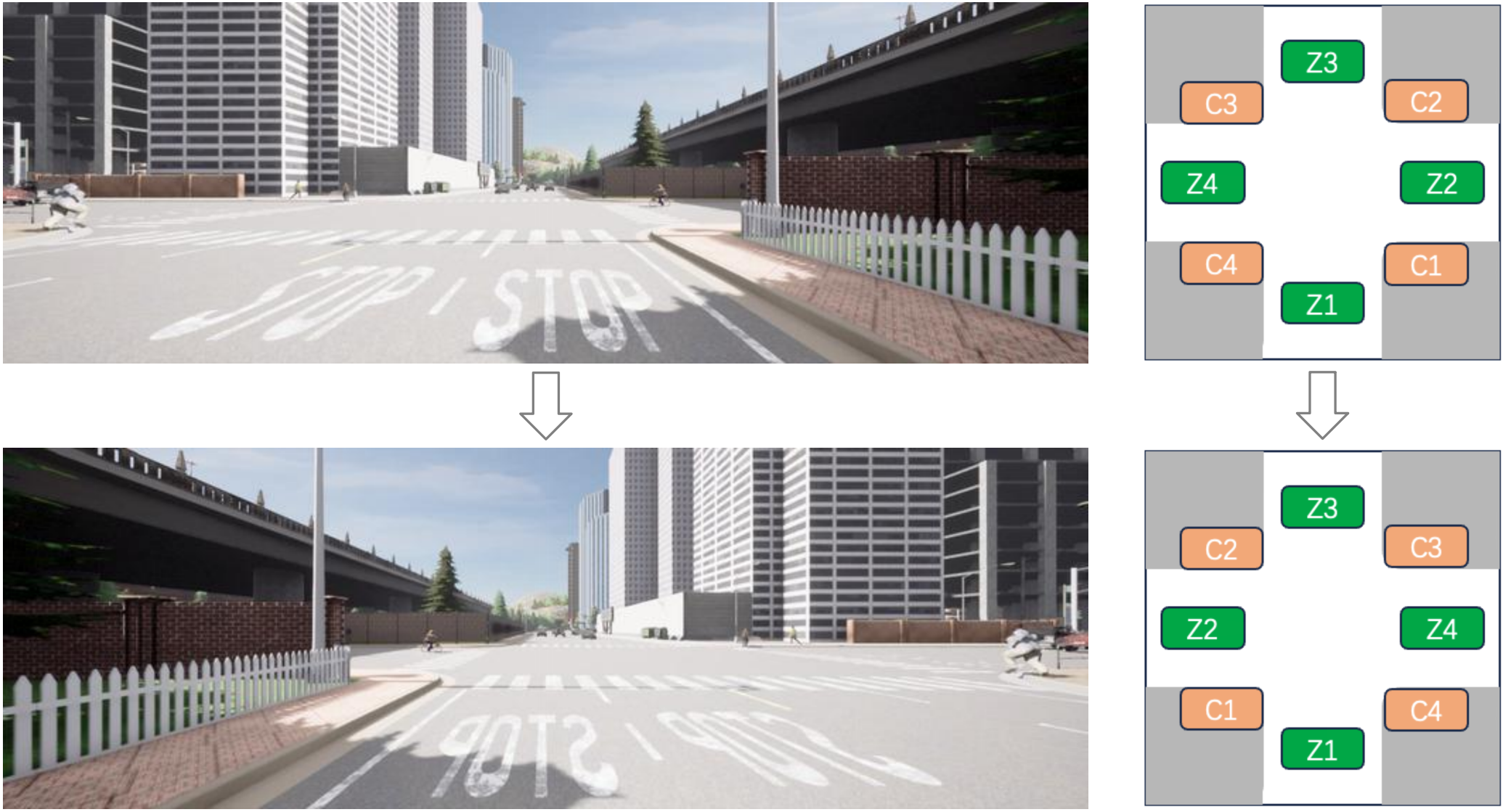}
  \caption{Video frame horizontal flipping.
  }
  \label{fig:video_flip}
\end{figure}

The second data augmentation method involves horizontal flipping of each video frame, as shown in~\cref{fig:video_flip}.
It is important to note that after performing this operation, the atomic activity categories of some objects are also flipped, hence we need to correct the classification labels.
As shown in the diagram on the right of~\cref{fig:video_flip}, we need to establish a mapping relationship between before and after the flipping operation, such as z1-z2 should be mapped to z1-z4, c1-c2 should be mapped to c4-c3, so that the original labels can be mapped to new labels. 
It should be noted that the labels between z1 and z3 do not need to be modified.

\subsection{Other Solutions}
\subsubsection{Upsampling}
We simply upsample the input images by a factor of two to address the previously mentioned issue of action recognition for small objects.
This brings two benefits, on one hand, the model can more easily capture information about small objects. 
On the other hand, the model can better recognize groups composed of adjacent closer objects.

\section{Experiments}
\label{sec:results}

\subsection{Implementation Details}
In our experiments, we utilize Action-slot~\cite{kung2023action} as the base model.
We train the model on the TACO dataset~\cite{kung2023action}.
Each model is trained for 100 epochs with AdamW~\cite{loshchilov2017decoupled} as the optimizer, a learning rate of 5e-4, weight decay of 0.07, and a batch size of 8.
We only apply Cutout and video frame flipping data augmentation with a probability of 0.5 during the first 50 training epochs.
It should be noted that when InternVideo is used as the backbone network, the frame sequence length is set to 16, and no up-sampling is performed on the input images.

\subsection{Results}

The results on the test set are shown in \cref{tab:headings}.
We achieve best performance on the test set by applying the proposed methods.

\begin{table}[htb]
  \caption{Results.
  }
  \label{tab:headings}
  \centering
  \begin{tabular}{l|ccccccc}
    \hline
    Method      & mAP           & mAP@C        & mAP@K        & mAP@P         & mAP@C+       & mAP@K+        & mAP@P+        \\ \hline
    Action-slot & 0.54          & 0.48         & 0.41         & 0.49          & 0.70         & 0.62          & 0.53          \\
    Ours        & \textbf{0.69} & \textbf{0.7} & \textbf{0.6} & \textbf{0.63} & \textbf{0.8} & \textbf{0.74} & \textbf{0.62} \\ \hline
    \end{tabular}
\end{table}

\section{Discussion}
Our solutions can be further improved.
Firstly, the computational efficiency of the model ensembling method is low. Therefore, how to implement these ensembling strgies in an end-to-end model should be explored in the future.
Besides, we believe that combining this task of atomic activity recognition with the popular Bird's Eye View (BEV) paradigm in autonomous driving could help the model better understand road structures and object behaviors. 
Due to lacking camera parameters and depth information, we have not implemented this idea. 
However, we believe it can be explored in the future.

\section{Conclusion}
This report focus on the multi-label atomic activity recognition task of the 2024 ECCV ROAD++ Challenge.
We first identify the challenges in this task, including small objects, discriminating between single object and a group of objects, as well as model overfitting.
A series of solutions are propsoed to address these challenges.
Firstly, we construct a multi-branch activity recognition framework that not only separates different object categories but also the tasks of single object and object group recognition, thereby enhancing recognition accuracy. 
Secondly, various model ensembling strategies are introduced, including integrations of multiple frame sampling sequences, different frame sampling sequence lengths, multiple training epochs, and different backbone networks.
Furthermore, we propose an atomic activity recognition data augmentation method, which greatly expands the sample space by flipping video frames and road topology, effectively mitigating model overfitting.
The proposed methods achieve best performance in the test set of Track 3 for the ROAD++ Challenge 2024.


%
%
\bibliographystyle{splncs04}
\bibliography{main}

\begin{thebibliography}{1}
\providecommand{\url}[1]{\texttt{#1}}
\providecommand{\urlprefix}{URL }
\providecommand{\doi}[1]{https://doi.org/#1}

\bibitem{devries2017improved}
DeVries, T.: Improved regularization of convolutional neural networks with cutout. arXiv preprint arXiv:1708.04552  (2017)

\bibitem{X3D}
Feichtenhofer, C.: X3d: Expanding architectures for efficient video recognition. In: 2020 IEEE/CVF Conference on Computer Vision and Pattern Recognition (CVPR). pp. 200--210 (2020). \doi{10.1109/CVPR42600.2020.00028}

\bibitem{kung2023action}
Kung, C.H., Lu, S.W., Tsai, Y.H., Chen, Y.T.: Action-slot: Visual action-centric representations for multi-label atomic activity recognition in traffic scenes (2024)

\bibitem{loshchilov2017decoupled}
Loshchilov, I.: Decoupled weight decay regularization. arXiv preprint arXiv:1711.05101  (2017)

\bibitem{singh2022road}
Singh, G., Akrigg, S., Di~Maio, M., Fontana, V., Alitappeh, R.J., Saha, S., Jeddisaravi, K., Yousefi, F., Culley, J., Nicholson, T., et~al.: Road: The road event awareness dataset for autonomous driving. IEEE Transactions on Pattern Analysis \& Machine Intelligence (01), ~1--1 (feb 5555). \doi{10.1109/TPAMI.2022.3150906}

\bibitem{wang2022internvideo}
Wang, Y., Li, K., Li, Y., He, Y., Huang, B., Zhao, Z., Zhang, H., Xu, J., Liu, Y., Wang, Z., Xing, S., Chen, G., Pan, J., Yu, J., Wang, Y., Wang, L., Qiao, Y.: Internvideo: General video foundation models via generative and discriminative learning. arXiv preprint arXiv:2212.03191  (2022)

\end{thebibliography}
\end{document}